\documentclass[10pt,twocolumn,letterpaper]{article}

\usepackage{iccv}
\usepackage{times}
\usepackage{epsfig}
\usepackage{graphicx}
\usepackage{amsmath}
\usepackage{amssymb}
\usepackage{amsfonts}
\usepackage{multirow}
\usepackage{booktabs}
\usepackage{xcolor}
\usepackage{fancyhdr}
\usepackage{comment}
\usepackage{enumitem}
\setlist{nolistsep}
\usepackage{subfigure}
\setlist[itemize]{noitemsep, topsep=0pt}
\usepackage{soul}
\usepackage[accsupp]{axessibility}

\usepackage[ruled,vlined,linesnumbered]{algorithm2e}
\usepackage{bm,color}
\usepackage{bbm}
\usepackage{caption}
\DeclareMathOperator*{\argmax}{arg\,max}

\SetKwProg{Func}{}{}{}

\usepackage[pagebackref=true,breaklinks=true,letterpaper=true,colorlinks,bookmarks=false]{hyperref}

\iccvfinalcopy 

\usepackage{xspace}

\def\iccvPaperID{8907} 

\ificcvfinal\fi

\begin{document}

\title{TARGET: Federated Class-Continual Learning via Exemplar-Free Distillation}

\author{Jie Zhang\thanks{Work done during internship at Sony AI.} \\
ETH Zurich\\
\texttt{zj.jayzhang@gmail.com} \\
\and
Chen Chen \& Weiming Zhuang \& Lingjuan Lyu\thanks{Corresponding author.} \\
Sony AI \\
\texttt{\{ChenA.Chen, weiming.zhuang, Lingjuan.Lv\}@sony.com} \\
}


\maketitle
\ificcvfinal\thispagestyle{empty}\fi

\begin{abstract}
This paper focuses on an under-explored yet important problem: Federated Class-Continual Learning (FCCL), where new classes are dynamically added in federated learning. 
Existing FCCL works suffer from various limitations, such as requiring additional datasets or storing the private data from previous tasks. In response, we first demonstrate that non-IID data exacerbates catastrophic forgetting issue in FL. Then we propose a novel method called TARGET (federat\textbf{T}ed cl\textbf{A}ss-continual lea\textbf{R}nin\textbf{G} via \textbf{E}xemplar-free dis\textbf{T}illation), which alleviates catastrophic forgetting in FCCL while preserving client data privacy. Our proposed method leverages the previously trained global model to transfer knowledge of old tasks to the current task at the model level. Moreover, a generator is trained to produce synthetic data to simulate the global distribution of data on each client at the data level. Compared to previous FCCL methods, TARGET does not require any additional datasets or storing real data from previous tasks, which makes it 
ideal for data-sensitive scenarios. 
\end{abstract}

\section{Introduction}

Federated Learning (FL) is a 
privacy-aware 
learning paradigm that facilitates 
collaborations among multiple entities (e.g., edge devices or organizations) ~\cite{fedavg,kairouz2021advances,li2020federated,zhao2018federated,hard2018federated}. Each entity or client in FL retains data locally and transfers only training updates to the central server for aggregation. 



Conventional FL studies assume that the data classes and domains are static, but the new classes could emerge and data domains could change over time in reality ~\cite{zhang2020class,mittal2021essentials,belouadah2019il2m,zhao2020maintaining}. For example, multiple health institutions could use FL to collaborate and train models to identify COVID-19~\cite{yang2020covid,ciotti2020covid,bjursell2020covid} strains; new COVID-19 strains, however, continue to emerge due to high mutation rate of virus. An intuitive solution to this issue of continuously emerging data classes is training new models from scratch, but this is impractical as it would require significant extra computation cost. Another method is transfer learning from the previously trained model, but this method suffers from catastrophic forgetting~\cite{kirkpatrick2017overcoming,kemker2018measuring,robins1995catastrophic,goodfellow2013empirical}, degrading performance on the previous classes.

To address these issues, recent research~\cite{CFLKD,FCIL,bui2018partitioned,jiang2021fedspeech} has introduced the concept of Continual Learning (CL)~\cite{zenke2017continual,parisi2019continual,thrun1998lifelong,shin2017continual,serra2018overcoming} within the FL framework. These methods, collectively referred to as Federated Continual Learning (FCL), aim to mitigate the problems of catastrophic forgetting in FL. 
In most FCL scenarios, new classes are dynamically added, which we call Federated Class-Continual Learning (FCCL). FCCL allows local clients to continuously collect new data, and new classes can be added at any time. 
Unfortunately, existing FCCL works 
suffer from various limitations. 
For example, Ma et al.~\cite{CFLKD} utilize an unlabeled surrogate dataset to address the catastrophic forgetting problem, which may be difficult to obtain in some data sensitive scenarios. Furthermore, the usage of an unlabeled surrogate dataset may not be ideal for certain types of data, as it may not capture the full complexity of the original data.
In CL, exemplar-based methods~\cite{rebuffi2017icarl,liu2021rmm,wang2022foster,douillard2020podnet} have achieved leading performance. An exemplar refers to a sample or instance of a previously seen data point that is retained in a memory buffer for future use in the learning process.
Dong et al.~\cite{FCIL} propose a exemplar-based method that stores historical data to address catastrophic forgetting. However, in many privacy-sensitive scenarios (\textit{e.g.}, hospitals and medical research institutions), users are not permitted to store data from previous tasks due to privacy and policy concerns and data will not be kept for a long time~\cite{kaissis2021end,vizitiu2019towards,liu2021data,smith2021always}. 
In summary, the majority of FCCL methods train the global model with additional datasets or previous task data, which could potentially violate data privacy regulations. This dilemma prompts us to consider the following question:

\textbf{\textit{Question:}} \textit{How to effectively alleviate the catastrophic forgetting problem in the FCCL without storing the local private data of the client or any additional datasets?}

To address this question, we conduct a systematic analysis and observe that the imbalanced distribution of data among clients in FL exacerbates the catastrophic forgetting problem (see Section~\ref{sec:global_info}). 
In order to fix this problem: 1) at the model level, we leverage the previously trained global model to transfer knowledge of the old tasks to the current task. 2) at the data level, we train a generator to produce synthetic data that aims to simulate the global distribution of data on each client. Drawing on these insights, we present a method called TARGET (federat\textbf{T}ed cl\textbf{A}ss-continual lea\textbf{R}nin\textbf{G} via \textbf{E}xemplar-free dis\textbf{T}illation) that mitigates catastrophic forgetting in FCCL without compromising clients' data privacy.

Our contributions can be concluded as follows:
\begin{itemize}
    \item We are the first to demonstrate that non-independent and identically distributed (non-IID) data exacerbates catastrophic forgetting issue in FL.
    Then we propose a novel method called TARGET, which alleviates the catastrophic forgetting in FCCL by leveraging global information.   
    \item Compared to previous FCCL methods, TARGET doesn't require extra datasets or data from previous tasks, it can be applied in data sensitive scenarios. 
    \item 
    Extensive experiments demonstrate the efficacy of our proposed method. For example, when partitioning the CIFAR-100 dataset into five tasks, our method achieves an accuracy of 36.31\%, which is about 6\% higher than the best baseline method.
\end{itemize}

\section{Related Work}
\subsection{Federated Continual Learning}
In recent years, deep neural networks (DNNs) have been widely used as a fundamental technology for the development of artificial intelligence, both in established and emerging fields~\cite{zhu2023universal,zhu2023generalized,li2023revisiting,edge_fl,9616392_Dong}. With the advancement of deep learning, an increasing number of researchers have started to focus on the construction of privacy-preserving deep learning frameworks. FL is a paradigm for collaboratively building a model across multiple clients \cite{konevcny2015federated,  mcmahan2017communication}, which is gaining momentum in recent years. However, 
Federated Continual Learning (FCL), which focused on both FL and CL simultaneously is just emerging and still remains pending further research \cite{usmanova2021distillation,park2021tackling,yoon2021federated}. Apart from the client-wise catastrophic forgetting, FCL paradigm also poses new challenges such as inter-client interference and communication-efficiency \cite{yoon2021federated}. FedWeIT solves these challenges through decomposing the parameters into three parts, i.e., global parameters, local based parameters, and task-adaptive parameters \cite{yoon2021federated}. Concurrently, Concept-Drift-Aware Federated Averaging (CDA-FedAvg) extends the popular FL algorithm, Federated Averaging (FedAvg), to tackle the CL problem by introducing the concept drift detection and adaptation \cite{estevez2021concept}. 
FCL with Distillation (CFeD) treats the model trained with the last task as the teacher model to perform a knowledge distillation and proposes a server distillation mechanism to deal with non-i.i.d. issue \cite{DBLP:conf/ijcai/MaX00S22}. Global-Local Forgetting Compensation (GLFC) designs a class-aware gradient compensation loss and a class-semantic relation distillation loss to prevent forgetting, and a proxy server to 
mitigate the non-independent and identically distributed (non-IID) 
problem \cite{dong2022federated}. Additionally, Federated Selective Inter-client Transfer (FedSeIT) applies FCL to NLP through selectively combining model parameters of foreign clients and selecting informative tasks to perform knowledge transfer \cite{chaudhary2022federated}. These papers illustrate the growing interest in FCL and the need for novel approaches to address its unique challenges. As research in this area continues to advance, we can expect to see more innovative approaches to FCL that further improve its performance, scalability, and privacy preservation. 

\subsection{Continual Learning}

Continual Learning has been studied extensively, several training methods have been proposed to address the catastrophic forgetting challenge it presents. Regularization-based approaches: Elastic Weight Consolidation (EWC) selectively penalizes the network parameters that are important for old tasks \cite{kirkpatrick2017overcoming}. Synaptic intelligence (SI) uses a memory buffer to store important network parameters \cite{zenke2017continual}. Incremental moment matching (IMM) modelings the posterior distribution after learning multiple tasks as a mixture of Gaussian models \cite{lee2017overcoming}. Stable SGD proposes to carefully design the training regimes such as learning rate decay, batch size, dropout, and optimizer to alleviate forgetting. Replay-based approaches: The methods in this family construct a memory to store the past information which will be presented to the model for reviewing in future tasks. Some store the knowledge of the previous tasks, known as experience replay \cite{lopez2017gradient,riemer2018learning,shin2017continual,rios2018closed}. iCaRL preserves the most representative samples of each class \cite{rebuffi2017icarl}. Averaged Gradient Episodic Memory (A-GEM) builds an episodic memory of model parameter gradients\cite{chaudhry2018efficient}. Achitecture-based: DEN expands the model size \cite{yoon2017lifelong} and RCL utilizes reinforcement learning \cite{xu2018reinforced}. APD divides the model parameters into shared and task-specific parameters to restrict the model complexity \cite{yoon2019scalable}.  

\paragraph{Exemplar-Free Continual Learning}
Besides, a promising line of work focuses on data-free Continual Learning. DeepDream perturbs current training samples into images that maximize "forgetting" from the previous tasks \cite{mordvintsev2015inceptionism}. DeepInversion proposes a model inversion technique and evaluates its performance in a Continual Learning scenario but found limited success \cite{yin2020dreaming}. DFCIL~\cite{smith2021always} investigates the failure and decomposes the CE-loss into two different losses which guarantee the learning of effective features. These methods enable Continual Learning in scenarios where storing old task data is not feasible or desirable, such as in privacy-sensitive applications.

\section{Catastrophic Forgetting in FCCL}
This section conducts an in-depth analysis of catastrophic forgetting problem in federated class-continual learning (FCCL). We start by providing a formal definition of the problem. 
Then, we investigate the forgetting issue in FCCL and discuss potential methods to mitigate it.

\subsection{Problem Definition}

Federated Class-Continual Learning (FCCL) focuses on the problem of learning models for new classes over time in FL.
An FCCL framework consists of a central server and multiple clients. All clients do not share their raw data with any other client or the central server.
Each client learns from a sequence of $n$ tasks, where $k$-th task contains non-overlapping subsets of classes $C_k\in C$ , where $C$ is the set of all possible classes.
In our privacy-aware scenario, the task stream is presented in an unknown order, and each client can only access its local data from task $k$ during that task's training period, which is no longer accessible thereafter.
Note that the models are trained in a distributed manner, where each party has access to only a subset of the classes $C_k$ (\ie non-IID).
We also consider a more challenging and practical setting where the data in each client is heterogeneous. 
In this paper, we assume the label distribution of data in each client is skewed~\cite{li2018federated,zhang2022federated}. 
\paragraph{Forgetting Issue in FCCL}
The goal of the global model optimization problem at task $k$ is to minimize the overall classification error on the current set of classes $C_k$. 
However, when a new task arises, clients are not able to access data from previous (old) tasks due to privacy concerns and can only update their local model with data from the new task. 
This often leads to a significant decrease in performance on previous tasks, which is known as catastrophic forgetting~\cite{kirkpatrick2017overcoming,kemker2018measuring,robins1995catastrophic,goodfellow2013empirical}.
To mitigate catastrophic forgetting in the global model, we aim to
minimize the overall classification error on the current set of classes $C_k$, while simultaneously minimizing the changes to the previously learned classes. Formally, the objective function can be written as:
\begin{equation}
\min_{\theta_k} \sum_{c \in C_k} \sum_{i=1}^{m_c} L(f_k(x_{i,c}; \theta_k), c) + \alpha R(\theta_k, \theta_{k-1})
\end{equation}
where $\theta_k$ is the model parameter at round $k$, $L$ is a loss function that measures the classification error, $R$ is a regularization term that penalizes changes to the previous model parameters, $m_c$ is the number of data in class $c$, and $\alpha$ is a hyper-parameter that controls the strength of the regularization. 
In this formula, $f_k(x_{i,c}; \theta_k)$ represents the classification model that takes as input a data point $x_{i,c}$ associated with class $c$ and outputs a probability distribution over the set of classes in $C_k$. 
The regularization term $R$ encourages the new model parameters to be close to the previous model parameters $\theta_{k-1}$, in order to prevent catastrophic forgetting of the previously learned classes.

\begin{figure}[t]
\centering
\subfigure[Test Accuracy]{
\begin{minipage}[t]{0.45\linewidth}
\centering
\includegraphics[width=1.5in]{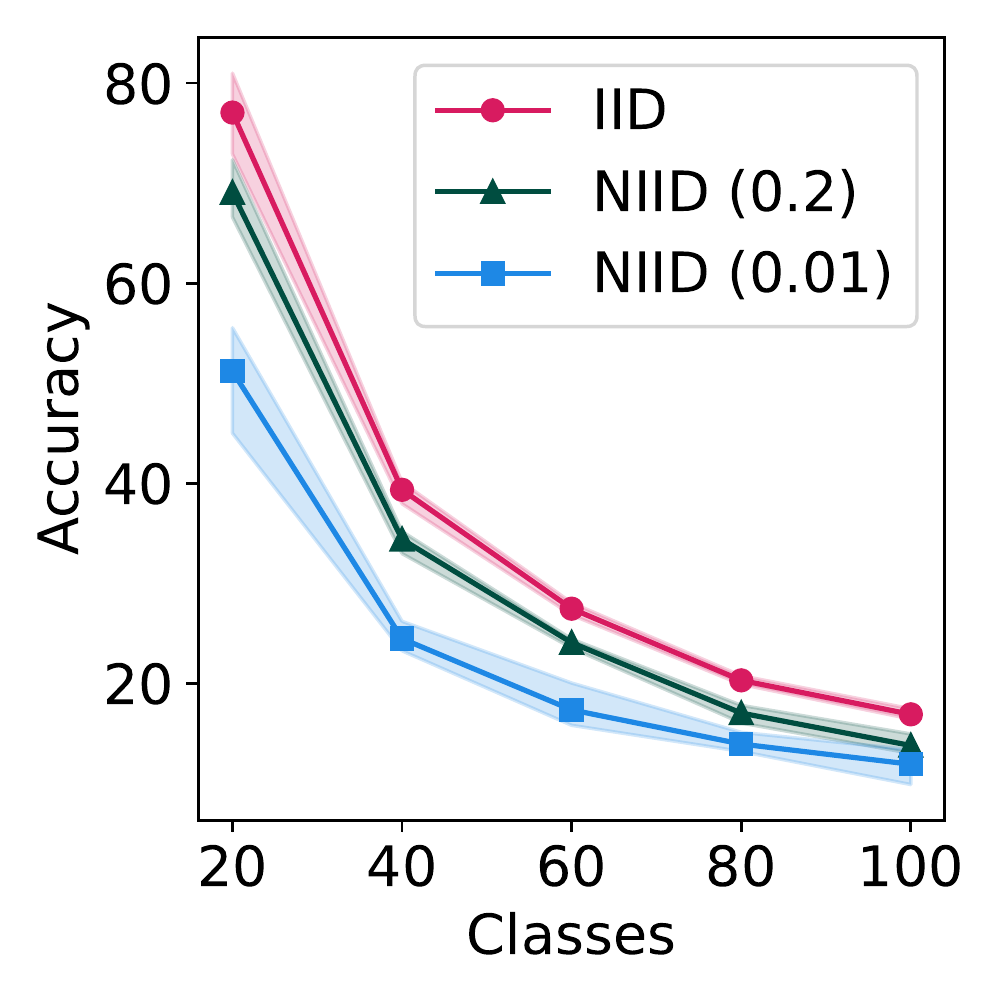}
\label{fig:forgetting_a}
\end{minipage}%
}%
\subfigure[Relative Forgetting]{
\begin{minipage}[t]{0.45\linewidth}
\centering
\includegraphics[width=1.5in]{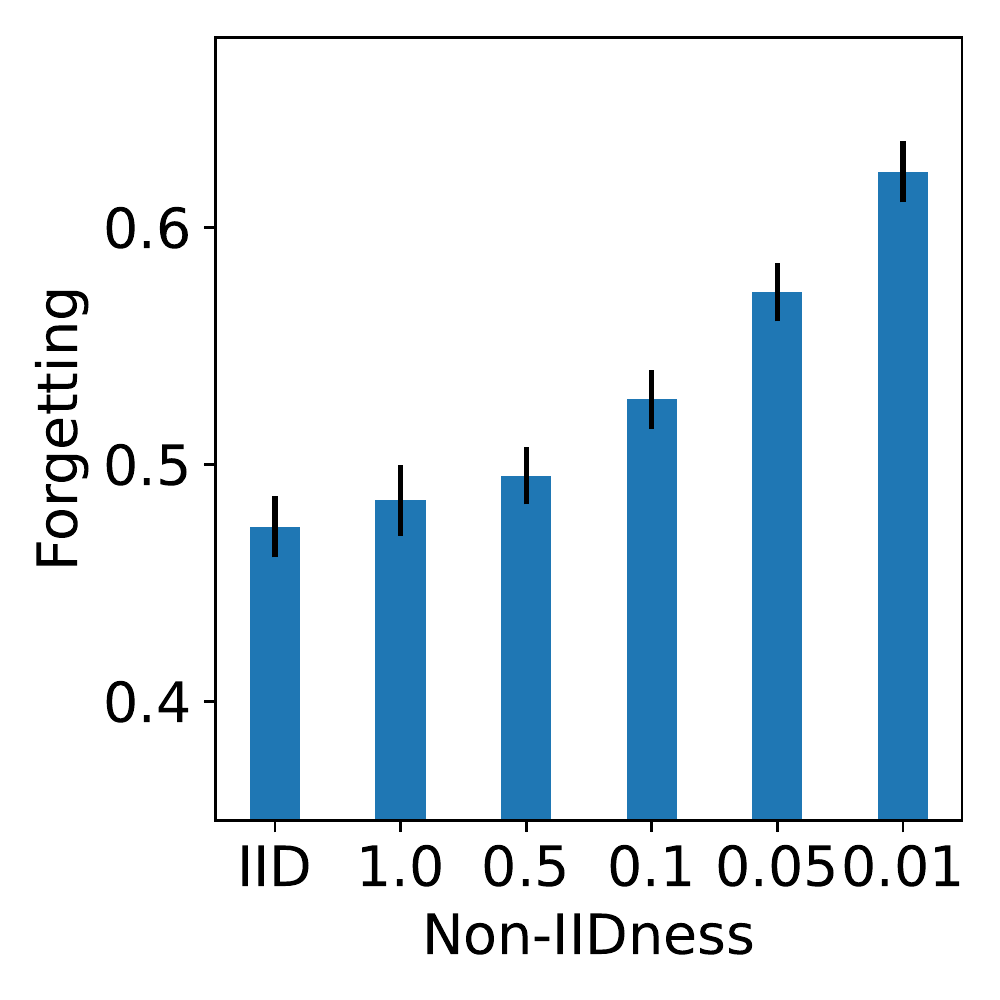}
\label{fig:forgetting_b}
\end{minipage}%
}%
\centering
\caption{Forgetting under different data partitions. We partitioned the CIFAR-100 dataset into five tasks with 20 classes each, distributed among five clients. ``NIID" refers to non-IID, where a lower value 
represents a more imbalanced or skewed distribution of data. Non-IIDness refers to the Dirichlet parameter.}
 \label{fig:forgetting}
\end{figure}

\subsection{Heterogeneous Data Exacerbates Forgetting}
We argue that the degree of data heterogeneity has a substantial impact on catastrophic forgetting. To verify this, we conduct an experiment on CIFAR-100 dataset~\cite{krizhevsky2009learning} with different degrees of data heterogeneity.
Inspired by Backward Transfer (BwT)~\cite{chaudhry2018riemannian}, we derive the following formula to measure the severity of forgetting, a popular forgetting measure in CL~\cite{cha2020cpr,chaudhry2020continual,chaudhry2018efficient}:
\begin{equation}
\mathcal{F}_k=\frac{1}{k-1} \sum_{j=1}^{k-1} f_j^k,
\label{eq:forget}
\end{equation}
where $\mathcal{F}_k$ denotes the average forgetting at $k$-th task and $f_j^k$ quantifies forgetting for the $j$-th ($j \textless k$) task after the model has been continually trained up to task $k$. Specifically, for a given data distribution, $f_j^k$ can be expressed as follows:
\begin{equation}
f_j^k=\frac{1}{\lvert \mathcal{C}^j \rvert } \sum_{c\in \mathcal{C}^j} \max _{t \in\{1, \ldots . {N}-1\}}\left(\mathcal{A}_c^{(n)}-\mathcal{A}_c^{({N})}\right),
\end{equation}
where $\mathcal{C}^j$ is a set of classes
related to the $j$-th task, $\mathcal{A}_c^{(n)}$ is the accuracy on class $c$ at round $t$, and $\mathcal{A}_c^{(N)}$ is the final accuracy on class $c$ after learning all tasks. Note that $f_j^k$ captures the average gap between the peak accuracy and the final accuracy for each class of the $j$-th task after learning the $k$-th task.

  


We further extend the catastrophic forgetting measurement $\mathcal{F}$ in Equation~\ref{eq:forget} to FL under different data partitions and introduce a relative metric $\mathcal{R}$ to measure forgetting as follows:
\begin{equation}
\mathcal{R}_k=\frac{\sum_{j=1}^{k-1} f_j^k}{ \sum_{j=1}^{k-1} \mathcal{A}_{(j,k)}},
\end{equation}
where $\mathcal{A}_{(j,k)}$ is the accuracy on task $j$ after learning task $k$. 
For different data partitions, an increased $\mathcal{R}_k$ indicates a more serious forgetting of previous tasks.

Figure~\ref{fig:forgetting} illustrates the impact of catastrophic forgetting under independent and identically distributed (IID) and different levels of non-independent and identically distributed (non-IID) data partitions. 
In particular, we employ the Dirichlet distribution, which is widely used in FL~\cite{li2018federated, li2020federated}, to simulate the imbalanced label distribution among different clients.
Figure~\ref{fig:forgetting_a} shows that the accuracy of the model degrades as training proceeds to new tasks in the IID setting. The performance is even worse in the non-IID settings. These results suggest that FCCL faces significant challenge on extreme non-IID settings. We further analyze the forgetting phenomenon in the Figure~\ref{fig:forgetting_b}. The higher degree of non-IID exacerbates the forgetting phenomenon in FCCL. These empirical studies motivate us to further investigate the catastrophic forgetting issue in 
FCCL. 

\subsection{Alleviating Forgetting via Global Information}
\label{sec:global_info}

To tackle the issue of catastrophic forgetting in the FCCL, we argue utilizing global information to improve performance. The global information can derive from the global model. We explore this approach and empirically demonstrate that the integration of global information can effectively mitigate catastrophic forgetting.


\paragraph{Learn from the Global Model}
Inspired by LwF~\cite{li2017learning}, we leverage the knowledge from the previously trained global model to the current task via Knowledge Distillation (KD), using only the data from the current task. 
We conduct experiments on CIFAR100 dataset with 5 continual tasks, where each task contains data of 20 classes. We evaluate the accuracy of the model on each previous task after training on all five tasks and report the final accuracy as the average of these results. Table~\ref{tb:global_model} shows that FCCL without knowledge from the global model (\textit{i.e.,} FedAvg~\cite{fedavg}) suffers severely from catastrophic forgetting under both IID and non-IID data partitions. Although FedAvg achieves competitive performance on Task 5, it results in 0\% accuracy on previously trained tasks (Task 1, 2, and 3). 
In contrast, FedLwF 
achieves better accuracy on Tasks 1, 2, 3, and 4 and final accuracy.
The experiments demonstrate that the global model can indeed alleviate the forgetting issue.



\begin{table}[t]
\caption{Test accuracy on previous tasks after training on all tasks. In this context, ``FedAvg" denotes a naive approach whereby clients learn tasks sequentially.}
\label{tb:global_model}
\scalebox{0.7}{
\centering
\begin{tabular}{cccccccc}
\toprule
Partition               & Method    & Task 1 & Task 2 & Task 3 & Task 4 & Task 5 & Final \\
\midrule
\multirow{2}{*}{IID}         & FedAvg    &    0    &    0    &    0    & 0.05       &   82.6     &   16.53    \\
                             & FedLwF &  6.8 \scriptsize{\textcolor{blue}{$\uparrow$}}    &   11.5 \scriptsize{\textcolor{blue}{$\uparrow$}}    &  27.1 \scriptsize{\textcolor{blue}{$\uparrow$}}     &   44.45 \scriptsize{\textcolor{blue}{$\uparrow$}}    &  63.2 \scriptsize{\textcolor{red}{$\downarrow$}}    &   30.61 \scriptsize{\textcolor{blue}{$\uparrow$}}   \\
\midrule
\multirow{2}{*}{NIID (1)}    & FedAvg    & 0      & 0      & 0      & 0      & 81.65  & 16.33 \\
                             & FedLwF & 6.65 \scriptsize{\textcolor{blue}{$\uparrow$}}   & 13.71 \scriptsize{\textcolor{blue}{$\uparrow$}}  & 29.60 \scriptsize{\textcolor{blue}{$\uparrow$}}  & 45.41 \scriptsize{\textcolor{blue}{$\uparrow$}} & 59.35 \scriptsize{\textcolor{red}{$\downarrow$}}  & 30.94 \scriptsize{\textcolor{blue}{$\uparrow$}} \\
 \midrule
\multirow{2}{*}{NIID (0.5)}  & FedAvg    & 0      & 0      & 0      & 0.15   & 77.30  & 15.49 \\
                             & FedLwF &   1.6  \scriptsize{\textcolor{blue}{$\uparrow$}}   &   10.65  \scriptsize{\textcolor{blue}{$\uparrow$}}   &   27.75  \scriptsize{\textcolor{blue}{$\uparrow$}}   &   41.3 \scriptsize{\textcolor{blue}{$\uparrow$}}    &   56.65  \scriptsize{\textcolor{red}{$\downarrow$}}    &   27.59  \scriptsize{\textcolor{blue}{$\uparrow$}}  \\
\midrule                             
\multirow{2}{*}{NIID (0.02)} & FedAvg    & 0      & 0      & 0      & 0.1    & 59.00  & 11.82 \\
                             & FedLwF &   0.1 \scriptsize{\textcolor{blue}{$\uparrow$}}    &     1.7 \scriptsize{\textcolor{blue}{$\uparrow$}}   &  3.35 \scriptsize{\textcolor{blue}{$\uparrow$}}     &   28.2 \scriptsize{\textcolor{blue}{$\uparrow$}}    &  54.0  \scriptsize{\textcolor{red}{$\downarrow$}}     &   17.47 \scriptsize{\textcolor{blue}{$\uparrow$}}  \\
\bottomrule

\end{tabular}}

\end{table}
\begin{figure}[t]
\centering
\includegraphics[width=8cm]{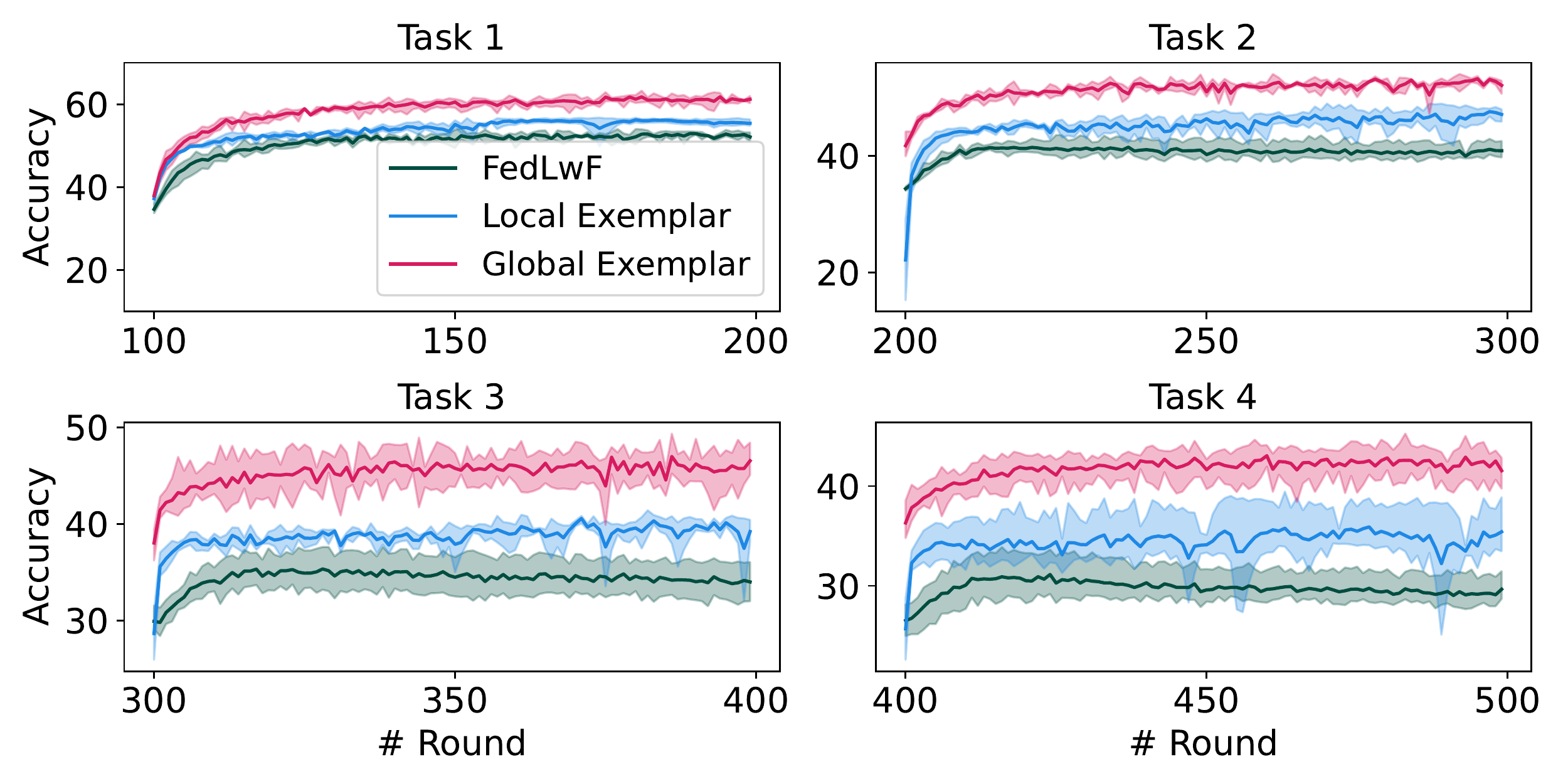}
\caption{Test accuracy by using current task's data, local exemplar and global exemplar.} 
\label{fig:global_local}
\end{figure}
\paragraph{Learn From Global Exemplar}
In continual learning, an exemplar refers to a sample from a previous task that is stored in a memory buffer for future training. 
Assuming that data stored in clients are compliant to use for future training, we adopt the idea proposed in iCaRL~\cite{rebuffi2017icarl} to save a small proportion of prior training data in memory. The selection of this small proportion of data assumes that the data distribution pertaining the old task is known, but this assumption does not hold in FL because clients are unable to know the data distributions of others due to data privacy concerns. Nevertheless, we employ two exemplar selection methods to illustrate the utility of global data: \textit{global exemplar} and \textit{local exemplar}. Global exemplar assumes that the server aggregates a subset of data from clients and distribute these data to clients in training (Note that this method does not conform to FL and is only used for comparison). Local exemplar means that each client retains a subset of data from local data in the previous tasks for future training.


As shown in Figure~\ref{fig:global_local}, we show the performance on the current and all previous tasks after learning the current task by using only the new task data, with local exemplar, and with global exemplar. It can be clearly observed that using exemplars can significantly improve the model performance, especially when using global exemplars, which can achieve much higher accuracy than using local exemplars. However, this raises a critical challenge of how to select such exemplars without violating the data privacy.

\section{Our Method: TARGET}

\subsection{Overview}
\begin{figure}[]
\centering
\includegraphics[width=8.6cm]{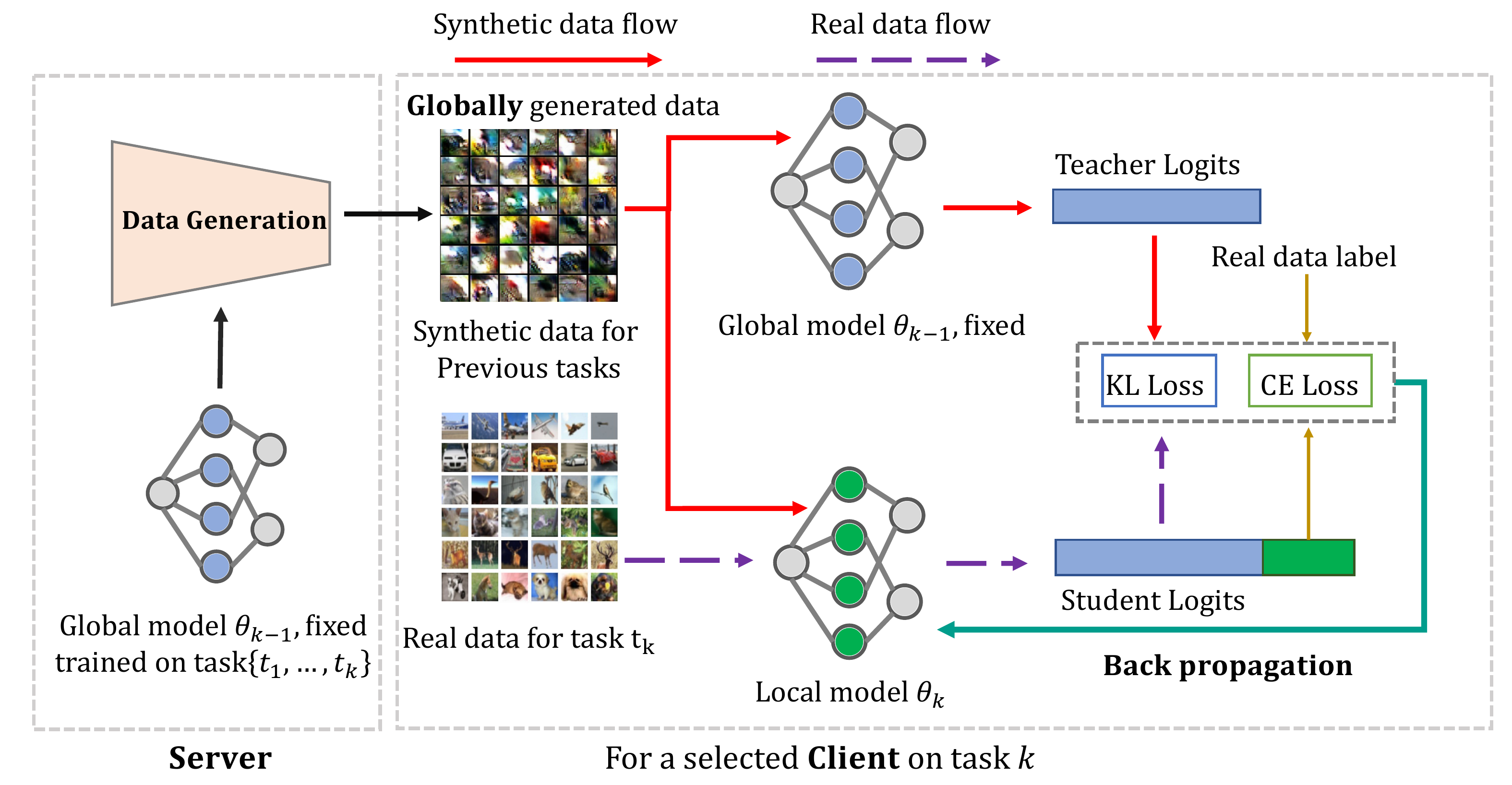}
\caption{Pipeline of TARGET. We utilize the global model (trained on task $k-1$) to synthesize the data with a global distribution, subsequently employing this synthesized data for training the $k$-th task.
}
\label{fig:pipeline}
\end{figure}
To 
utilize the global information without touching on the real exemplars from clients, we present a method called TARGET (federat\textbf{T}ed cl\textbf{A}ss-continual lea\textbf{R}nin\textbf{G} via \textbf{E}xemplar-free dis\textbf{T}illation), which utilizes global information without storing any real data. 
A detailed procedure is provided in Algorithm~\ref{alg:algorithm}.  Figure~\ref{fig:pipeline} presents an illustration of TARGET, wherein we synthesize data by inverting the global model $\theta_{k-1}$ (which was trained on task $k-1$), followed by combining the synthesized data with real data for local model update on task $k$.

\begin{algorithm}[t]
\caption{Procedure of TARGET.}
\label{alg:algorithm}
\KwIn{$B$: local minibatch size,
$E$: local epochs, $\eta$: learning rate, synthetic data: $X_{syn}=\emptyset$.}
\BlankLine
\ForEach{each task $\tau=0, 1, \dots$}{
    Initialize $w_0$ \\
    \ForEach{each round $t = 1, 2, \dots$}{
      $S_t \leftarrow$ (random set of $m$ clients) \\
      \ForEach{each client $k \in S_t$ {in parallel}}{
      $w_{t+1}^k \leftarrow \text{\textbf{ClientUpdate}}(k, w_t, \tau, X_{syn})$ 
      }
      $w_{t+1} \leftarrow \sum_{k=1}^m \frac{n_k}{n} w_{t+1}^k$
      
    }
    $X_{syn}=\text{\textbf{DataGeneration}}(w_{t+1})$ 

}
\Func{\textbf{ClientUpdate}($k, w, \tau, X_{syn}$):}{
    $\mathcal{B} \leftarrow$ (split  $\mathcal{P}_k \cup X_{syn}$ into batches) \\
    set global model $\mathcal{T} \leftarrow w$ \\
    \ForEach{each local epoch $i$ from $1$ to $E$}{
        \ForEach{batch $(b, b_{syn}) \in \mathcal{B}$}{
        $\ell(w; b)=L_{ce}(w;b)$ \\
        \If{$\tau \neq 0$}{
        $\ell(w; b) +=\alpha L_{kl}(w,\mathcal{T};b_{syn}) $
        }
        }
    }
    return $w$ to server
}
\Func{\textbf{DataGeneration}($w$):}{
Initialize parameter $\boldsymbol{\theta}_G, \boldsymbol{\theta}_S, X_{syn}=\emptyset$ \\
  \ForEach{round $i = 1, 2, \dots$}{
    Sample noises and labels $\{\mathbf{z}_i, \mathbf{y}_i\}_{i=1}^b$ \\
    \tcp{{data generation stage} }
        \ForEach{$j=1,2,\dots,$}{
        Generate $\{\hat{\mathbf{x}}_i\}_{i=1}^b$ with $\{\mathbf{z}_i\}_{i=1}^b$ and $G(\cdot)$ \\
        Update $\boldsymbol{\theta}_G$ using Equation~\ref{eq:generator}
        }
        Add a batch of data into $X_{syn}$ \\
        \tcp{{model distillation stage}} 
        Update $\boldsymbol{\theta}_S$ using Equation~\ref{eq:kd}
    }
    return $X_{syn}$ 

}
\end{algorithm}

\subsection{Server Side: Synthesizing Data for Old Tasks}
As demonstrated in Figure~\ref{fig:global_local}, data with global distributional information is more effective in mitigating the problem of catastrophic forgetting. Therefore, we propose a method of synthesizing data that can model the data distribution of the global model, without the need to preserve any client's privacy data. Specifically, given a global (teacher) model $\theta_{k-1}$ trained on task $k-1$, we first initialize a generator $G$ and a student model $\theta_{S}$ . We then repeatedly perform the following two training steps (see line 19$\sim$28 in Algorithm~\ref{alg:algorithm}): 1) update the generator by continuously optimizing it to generate data that conforms to the global model distribution; 2) update the student model by distilling knowledge from the teacher model with the synthetic data, hoping that the student model can learn the knowledge of the teacher model sufficiently, which demonstrates the effectiveness of the synthesized data.

\paragraph{Data Generation}



First, we utilize $G$ to generate synthetic data from noise $z$, we need to ensure that the synthetic data $\hat{x}=G(z)$ is similar to the training dataset. If the synthetic data is similar to the training dataset, their predictions should also be similar. We minimize the cross-entropy (CE) loss on the output of global model $\theta_{k-1}(\hat{x})$ and random labels $\hat{y}$,
\begin{equation}
    \mathcal{L}_G^{ce}=CE(\theta_{k-1}(\hat{x}), \hat{y}).
\end{equation}

It is expected that the synthetic data generated by generator can be classified into a particular class with a high degree of confidence. However, utilizing only the CE loss will cause the generator overfitting to the synthetic data that are far away from the decision boundary (of the global model)~\cite{zhangdense,fang2022up}, thus 
failing to deliver a good performance. 
In order to generate samples that are closer to the decision boundary (of the global model) with better transferability, following previous work~\cite{zhangdense}, we introduce a boundary support loss. Additional weight is given to the data 
on which the global model and the student model diverge in decision making.
\begin{align}
    &\mathcal{L}_G^{div}=-\omega KL(\theta_{k-1}(\hat{x}),\theta_S(\hat{x})), \text{and }\\
    &  \omega=\mathbbm{1}(\argmax\theta_{k-1}(\hat{x})\neq \argmax \theta_S(\hat{x})),
    \label{eq:student}
\end{align}
where $KL$ denotes the Kullback-Leibler (KL) divergence loss, $\mathbbm{1}(a)$ output 1 if $a$ is true and output 0 if $a$ is false. By maximizing the KL divergence loss, the generator can generate more representative data.

Motivated by~\cite{yin2020dreaming}, in order to further improve the stability of generator training, we introduce Batch Normalization (BN) loss to make synthetic data conform with the batch normalization statistics.
\begin{equation}
    \mathcal{L}_G^{bn}=\sum_l(||\mu_l(\hat{x})-\mu_{l}||+||\sigma_l^2(\hat{x})-\sigma_{l}^2||),
\end{equation}
where $\mu_l(\hat{x})$ and $\sigma_{l}^{2}(\hat{x})$ are the batch-wise mean and variance estimate of the $l$-th BN layer of the generator, $\mu_{l}$ and $\sigma_{l}^2$ are the mean and variance of the $l$-th BN layer of $f_S(\cdot)$. 

Combining the above losses, we can obtain the loss of the generator as follows,
\begin{equation}
\mathcal{L}_G = \mathcal{L}_G^{ce} + \lambda_1 \mathcal{L}_G^{div} + \lambda_2 \mathcal{L}_G^{bn},
\label{eq:generator}
\end{equation}
where $\lambda_1$ and $\lambda_2$ is the weight for different loss functions.

\paragraph{Model Distillation} 
In Equation~\ref{eq:student}, we introduce a student model to assist in training the generator to produce data with greater diversity. A better student model should lead to a better generator. Therefore, after training the generator for several rounds, we subsequently train the student model using the saved synthesized data and the output of the teacher model, using KL loss for knowledge distillation:
\begin{equation}
    \mathcal{L}_{S}=KL(\theta_{k-1}(\hat{x}),\theta_S(\hat{x})).
    \label{eq:kd}
\end{equation}
In this way, we can train a student model with better performance, and then further use it to update the generator. An ideal synthetic dataset should be able to efficiently enable student $\theta_S$ to fully learn the knowledge of teacher model.

Note that when the training of the whole process is over (\ie the student model can use the synthesized data to obtain high performance), we only retain the synthetic dataset $X_{syn}$ and transfer it to the clients, without saving the generator and student model.


\subsection{Client Side: Update with Global Information}
On the client side, we can obtain the data synthesized for the previous task $X_{syn}$ and the real training data of the current task $X_{local}$, then we train the local model $\theta_k$ for task $k$ on the two datasets at the same time. We showed in Section~\ref{sec:global_info} that the use of global models and global data can alleviate forgetting. Thus we distill the knowledge of global teacher model and global synthetic data by minimizing the following objective function,
\begin{equation}
    \mathcal{L}_{client} = \underbrace{CE(\theta_{k}({x}), {y})}_{\text{for current task}} + \alpha \cdot \underbrace{KL(\theta_{k-1}(\hat{x}), \theta_{k}(\hat{x}))}_{\text{for previous tasks}},
\end{equation}
where $(x,y)\in{X_{local}}$ and $(\hat{x})\in{X_{syn}}$. 
%
The utilization of the distillation loss facilitates efficient transfer of knowledge from the previous task to the current task model. And $\alpha$ is a hyper-parameter that controls the strength of the regularization for the previous tasks.
\section{Experiments}
\subsection{Experimental Settings}
We experiment on two datasets, namely
CIFAR-100~\cite{krizhevsky2009learning}, and Tiny-ImageNet~\cite{le2015tiny}
, to evaluate the performance of our proposed approach. 
To establish the order of the continual tasks, we adopt the widely used protocols~\cite{shin2017continual,zenke2017continual,parisi2019continual,chaudhry2020continual}. Specifically, we divide all classes of each dataset equally into 
multiple tasks by default, \ie we evenly divide the classes into 5 and 10 tasks to simulate class continual 
learning scenarios. 
We employ ResNet18~\cite{he2016deep} as the backbone for the classification model. 

\begin{table*}[t]
\centering
\caption{The Average Accuracy (\%) and Forgetting for all learned tasks on CIFAR-100 for various numbers of tasks (5, 10) under both IID and non-IID settings. Results are reported as an average of 3 runs. 'Acc' refers to average accuracy, and '$\mathcal{F}$' represents the forgetting measure utilized in Equation~\ref{eq:forget}. The best results are in bold.}
\label{tb:main}
\begin{tabular}{c|cc|cc|cc|cc|cc|cc}
\toprule
Data partition & \multicolumn{4}{c|}{IID}     & \multicolumn{4}{c|}{NIID (1)} & \multicolumn{4}{c}{NIID (0.5)} \\
\midrule
Tasks & \multicolumn{2}{c|}{T=5} & \multicolumn{2}{c|}{T=10} & \multicolumn{2}{c|}{T=5} & \multicolumn{2}{c|}{T=10} & \multicolumn{2}{c|}{T=5} & \multicolumn{2}{c}{T=10} \\
\midrule
Method         & Acc($\uparrow$)   & $\mathcal{F}$($\downarrow$)    & Acc($\uparrow$)   & $\mathcal{F}$($\downarrow$)    &Acc($\uparrow$)   & $\mathcal{F}$($\downarrow$)    &
Acc($\uparrow$)   & $\mathcal{F}$($\downarrow$)    &Acc($\uparrow$)   & $\mathcal{F}$($\downarrow$)    &  Acc($\uparrow$)   & $\mathcal{F}$($\downarrow$)    \\
\midrule
Finetune         & 16.12 & 0.78 & 7.83  & 0.75 & 16.33  & 0.77 & 8.45  & 0.74 & 15.49  & 0.74  & 7.64   & 0.71 \\
\midrule
FedEWC         & 16.51 & 0.71 & 8.01  & 0.65 & 16.06  & 0.68 & 8.84  & 0.62 & 16.86  & 0.66  & 8.04   & 0.65 \\
\midrule
FedWeIT        & 28.45 & 0.52 & 20.39 & 0.43 & 28.56  & 0.49 & 19.68 & 0.45 & 24.57  & 0.54  & 15.45  & 0.48 \\
\midrule
FedLwF         & 30.61 & 0.45 & 23.27 & 0.37 & 30.94  & 0.42 & 21.16 & 0.41 & 27.59  & 0.44  & 17.98  & 0.45 \\
\midrule
Ours &
  \textbf{36.31} &
  \textbf{0.22} &
  \textbf{24.76} &
  \textbf{0.26} &
  \textbf{34.89} &
  \textbf{0.24} &
  \textbf{22.85} &
  \textbf{0.26} &
  \textbf{33.33} &
  \textbf{0.27} &
  \textbf{20.71} &
  \textbf{0.29} \\
  \bottomrule
\end{tabular}
\end{table*}

To evaluate our approach, we employ the standard continual learning metrics, as used in prior works~\cite{kirkpatrick2017overcoming,kemker2018measuring,robins1995catastrophic,goodfellow2013empirical}, which include \textbf{average accuracy} across all tasks and a \textbf{forgetting measure}~\cite{chaudhry2018riemannian} (see Equation~\ref{eq:forget}). For a fair comparison with the baseline class continual learning
methods in the FCCL setting, we implement three types of baselines: 1) Finetune, 
in which each client
simply learns tasks in sequence;
2) FedWeIT~\cite{yoon2021federated}, a regularization-based method in Federated Continual Learning that maximizes the knowledge transfer between
clients; 3) Examples of typical continual learning methods that do not store training data for rehearsal, including EWC~\cite{lee2017overcoming} and LwF~\cite{li2017learning}. In addition, we compare our method with methods that store real training data of old tasks, such as iCaRL~\cite{rebuffi2017icarl}. We implement these traditional continual learning algorithms in the FCCL scenario and name them as FedEWC, FedLwF, and FedIcaRL. For detailed information on the task configuration, default hyper-parameters and additional experimental results, please refer to the \textbf{Appendix}. 



\subsection{Experiments on CIFAR-100}
For CIFAR-100 dataset, we conduct experiments on two sets of tasks consisting of 5 and 10 tasks, respectively. 
We run the experiment on both IID and non-IID scenarios.
For non-IID setting, the Dirichlet parameter is set to $0.5$ and $1$, \ie NIID(0.5) and NIID(1).
Table~\ref{tb:main} 
shows the final average accuracy of the FL model trained on all tasks, along with the corresponding forgetting measure for each experiment. It is important to emphasize that an optimal method is characterized by 
high average accuracy and low forgetting measure. 

\begin{figure}[t]
\centering
\includegraphics[width=8.5cm]{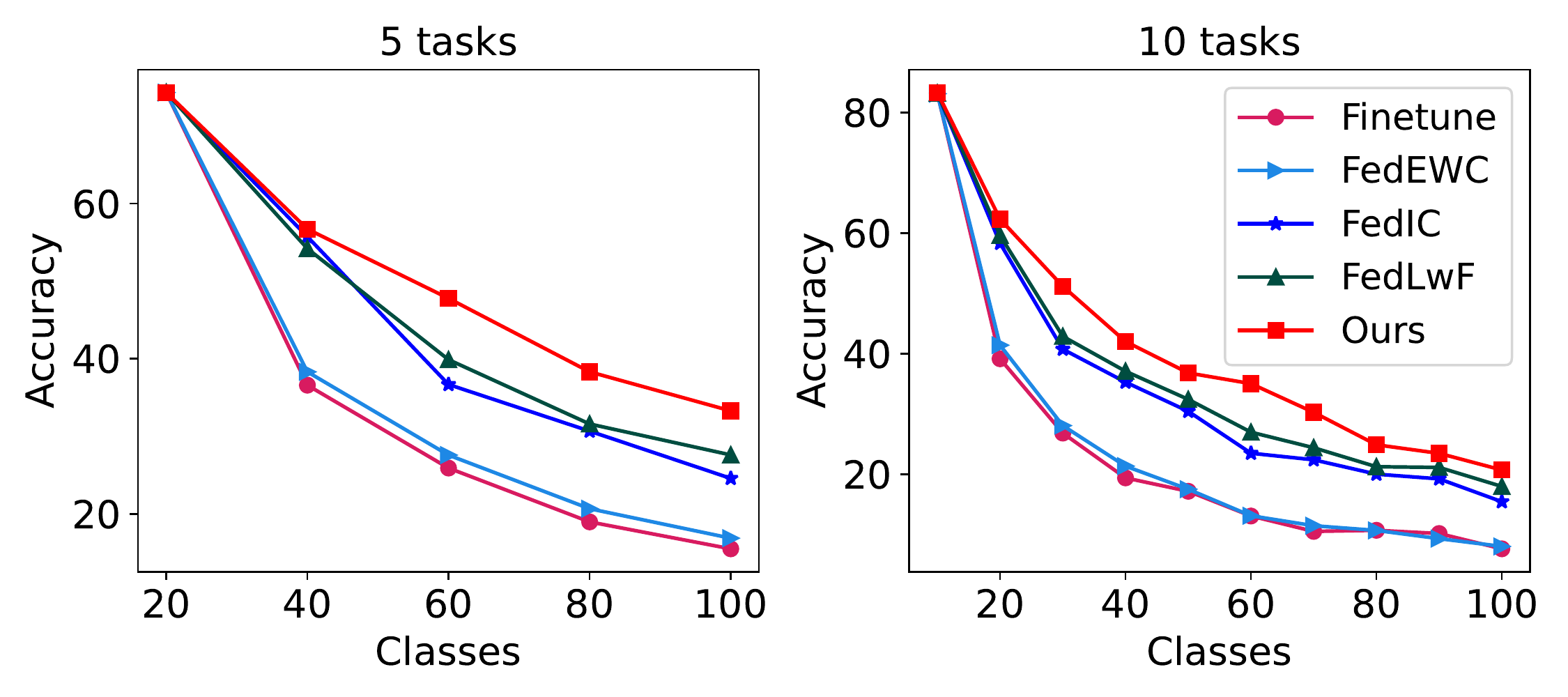}
\caption{Average accuracy on previous tasks and current task after the model trained on current task. }
\label{fig:5_10_cifar100_acc}
\end{figure}

Table~\ref{tb:main} indicates that Finetune exhibits the poorest performance when attempting to learn the continuously incoming tasks sequentially, thereby experiencing the issue of catastrophic forgetting. Moreover, we observed that methods based on regularization constraints, such as FedEWC and FedWeIT, were often ineffective in preventing the model from forgetting old tasks due to the lack of available data. On the other hand, we found that distillation-based approaches such as FedLwF and our proposed method were capable of improving the final average accuracy while simultaneously mitigating the issue of catastrophic forgetting. This is due to the transfer of model knowledge learned from the old task to the new task in a distillation manner, enabling the model to prevent catastrophic forgetting. Obviously, we found that applying our proposed method to generate synthetic datasets for the federated models trained on old tasks and subsequently performing model distillation on these synthetic datasets can lead to a substantial improvement in the average accuracy and reduction in the forgetting measure. This observation underscores the ability of our synthetic data to capture the distribution characteristics of historical task data accurately.
For example, in IID setting, when partitioning the CIFAR-100 dataset into five tasks, our method achieves an accuracy of 36.31\%, which is about 6\% higher than the best baseline method FedLwF. It is worth noting that we observed a decrease in the average accuracy of all methods to varying degrees as the number of tasks increased from 5 tasks to 10 tasks due to increased task complexity and the associated forgetting phenomenon. Nonetheless, our proposed method maintains the highest average accuracy and the lowest forgetting measure even under these more demanding conditions.

\begin{figure}[t]
\centering
\includegraphics[width=8.5cm]{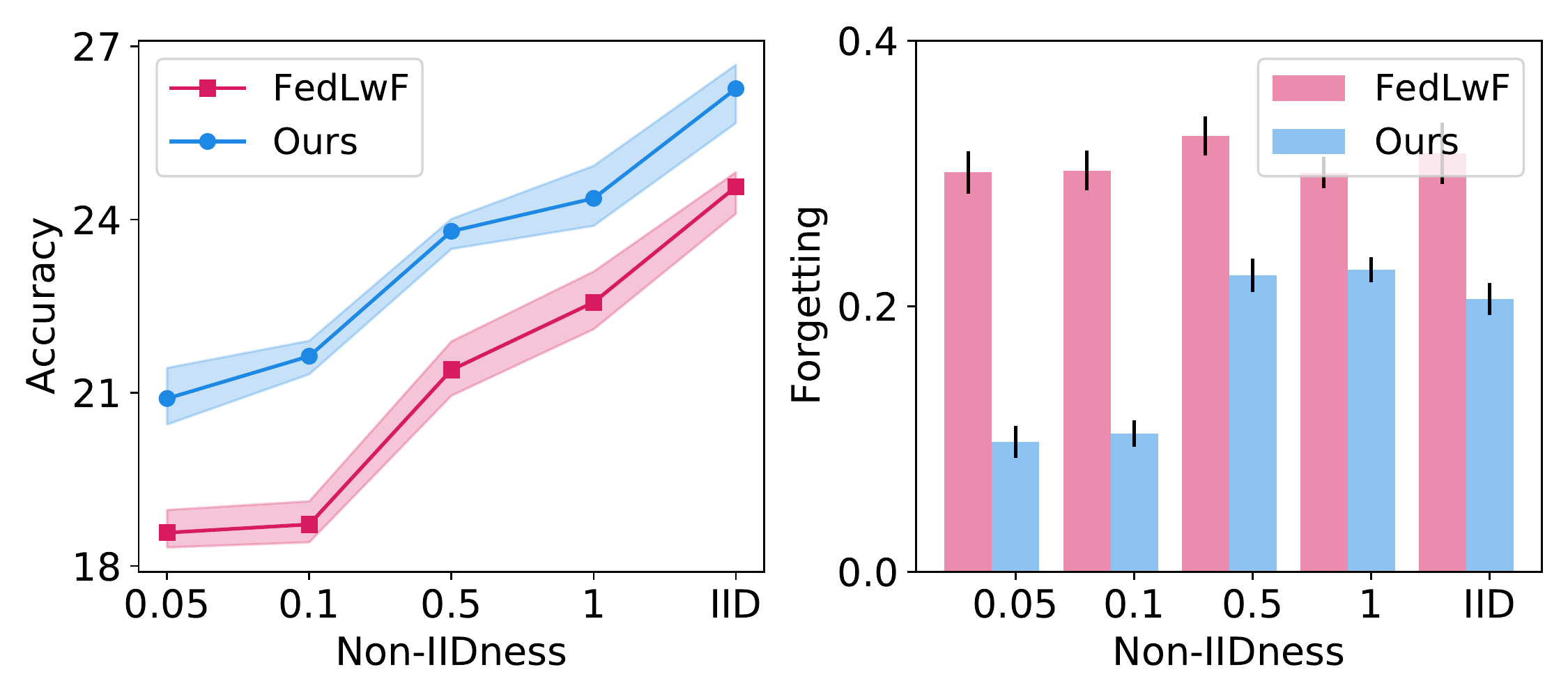}
\caption{The average accuracy (\%) and Forgetting for all learned tasks on Tiny-ImageNet for 5 tasks under both IID and non-IID settings.} 
\label{fig:tiny_acc}
\end{figure}

Figure~\ref{fig:5_10_cifar100_acc} illustrates the performance of the models trained by various methods on all previously learned tasks after the completion of each task. Specifically, it shows the average accuracy of the model on both the current and previous tasks after the completion of each task (e.g., after learning the second task, the average accuracy of the model on both the first and second tasks is measured). Based on these curves, it is evident that our proposed model outperforms other competing baseline methods in all incremental tasks, regardless of the number of tasks involved. This finding underscores the effectiveness of our approach in facilitating multiple local clients to learn new classes in a streaming manner while mitigating the forgetting problem.

\subsection{Experiments on Tiny-ImageNet}
We also evaluated the performance of our proposed method on the more challenging Tiny-ImageNet dataset and obtained similar results to those observed in the CIFAR-100 experiments. Specifically, in Figure~\ref{fig:tiny_acc}, we present the final average accuracy and forgetting measure for all learned tasks in both IID and non-IID settings for the case of 5 tasks. The Dirichlet parameter  is set to $\{0.05, 0.1, 0.5, 1\}$. Based on Figure~\ref{fig:tiny_acc}, it is evident that our proposed method consistently outperforms FedLwF in terms of average accuracy across all data partitions. Moreover, our method demonstrates a significantly lower forgetting measure than FedLwF under both IID and non-IID settings. Our proposed method achieves an average accuracy that is approximately 3\% higher than that of FedLwF even in the most challenging scenario (\ie NIID(0.05)). 
This result highlights the effectiveness of our method in mitigating catastrophic forgetting in the presence of extreme data distributions.

\subsection{Comparison with Exemplar-based Method}
\begin{figure}[t]
\centering
\includegraphics[width=8.8cm]{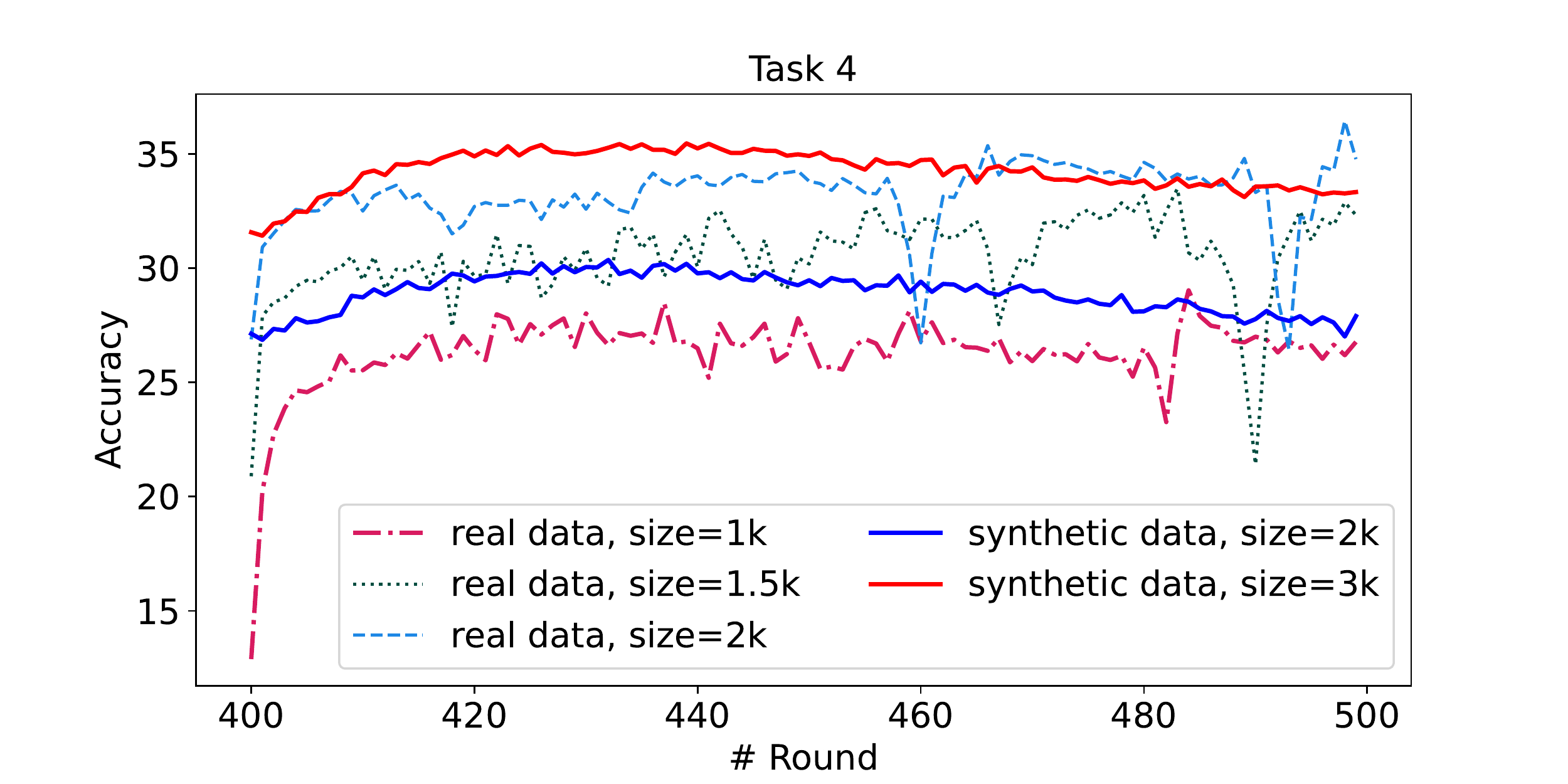}
\caption{Accuracy comparison 
between real data-based method (iCaRL) and synthetic data-based method (TARGET). We show the result for the last task.}
\label{fig:real_synthetic}
\end{figure}

In CL, the most successful approaches to alleviate forgetting require extensive replay of previously seen data, which can be problematic when data legality and privacy concerns exist. 
Among them, iCaRL~\cite{rebuffi2017icarl} is a classic but unrealistic algorithm that relies on the stored exemplars in addition to the network parameters, and it is intuitive that using  old task's real data could be beneficial to alleviate forgetting problem. 
To further understand the performance gap, we compare our proposed TARGET (which uses synthetic data) with iCaRL (which 
requires storing exemplars from old task's real data), 
and study the effects of different exemplar memories on the performance of our method and iCaRL in Figure~\ref{fig:real_synthetic}. We set the stored exemplar size to \{1000, 1500, 2000\} for iCaRL, and \{2000, 3000\} for our method on the CIFAR-100 dataset. 

The accuracy curve in Figure~\ref{fig:real_synthetic} represents the average accuracy rate measured over all 100 classes learned by the model during the last learning task. While our method can achieve similar performance to storing 1k real training data by storing 2k synthetic data, it still cannot outperform storing 2k real training data. 
However, it is worth noting that our method does not require storing any real training data, which can be a significant advantage in scenarios where storing real data is difficult or not allowed due to privacy or legal concerns. Additionally, our method achieves better performance than storing only 1k real training data, which indicates that our synthetic data is effective in mitigating catastrophic forgetting. We observed that when our method stores 3k synthetic data, it achieves better accuracy than when it stores 2k synthetic data. However, surpassing iCaRL in performance with an equal amount of data remains a challenge for our method. 
How to effectively use fewer synthetic data 
with more valuable knowledge from previous tasks will be left as a future research direction.
\subsection{Analysis of Our Method}
\begin{table}[t]
\centering
\caption{ The effect of $\alpha$ on the performance of both new and old tasks in CIFAR-100, 2 tasks. }
\label{tb:kd}
\scalebox{0.9}{
\begin{tabular}{c|ccccc}
\toprule
$\alpha$         & 3     & 5     & 10             & 15             & 25    \\
\midrule
Old Task (0-49)   & 28.34 & 34.91 & 49.02          & 58.48          & 66.02 \\
\midrule
New Task (50-99)  & 62.46 & 60.26 & 53.74         & 45.26          & 29.32 \\
\midrule
Average (0-99) & 45.41 & 47.58 & \underline{51.38} & \textbf{51.87} & 47.76 \\
\bottomrule
\end{tabular}}
\label{tb:kd_weight}
\end{table}
\paragraph{Trade-off between Backward and Forward Transfer.}
Continual learning presents a challenge in balancing the trade-off between maintaining high accuracy on old tasks (backwards transfer) and achieving high accuracy on new tasks (forward transfer). In Table~\ref{tb:kd_weight}, we evenly split the CIFAR-100 dataset into two tasks and test the average accuracy of our proposed method under different values of $\alpha$. We observe that the trade-off between backward and forward transfer is not always balanced. In order to achieve good backward transfer, a large value of $\alpha$ should be used to prevent the model from forgetting previous tasks and to encourage it to focus more on the old synthetic data. Conversely, to achieve good forward transfer, a small value of $\alpha$ should be used, allowing the model to learn quickly and effectively from new tasks while still maintaining some knowledge from the old tasks. The experimental results in Table~\ref{tb:kd_weight} indicates that when the value of $\alpha$ is between 10-15, the model achieves a good balance between accuracy on new tasks and accuracy on old tasks. We alspo partition the CIFAR-100 dataset into 5 equal tasks and test our method's performance on all previous tasks after learning each new task (refer to {Appendix}). 

\begin{figure}[t]
\centering
\includegraphics[width=8cm]{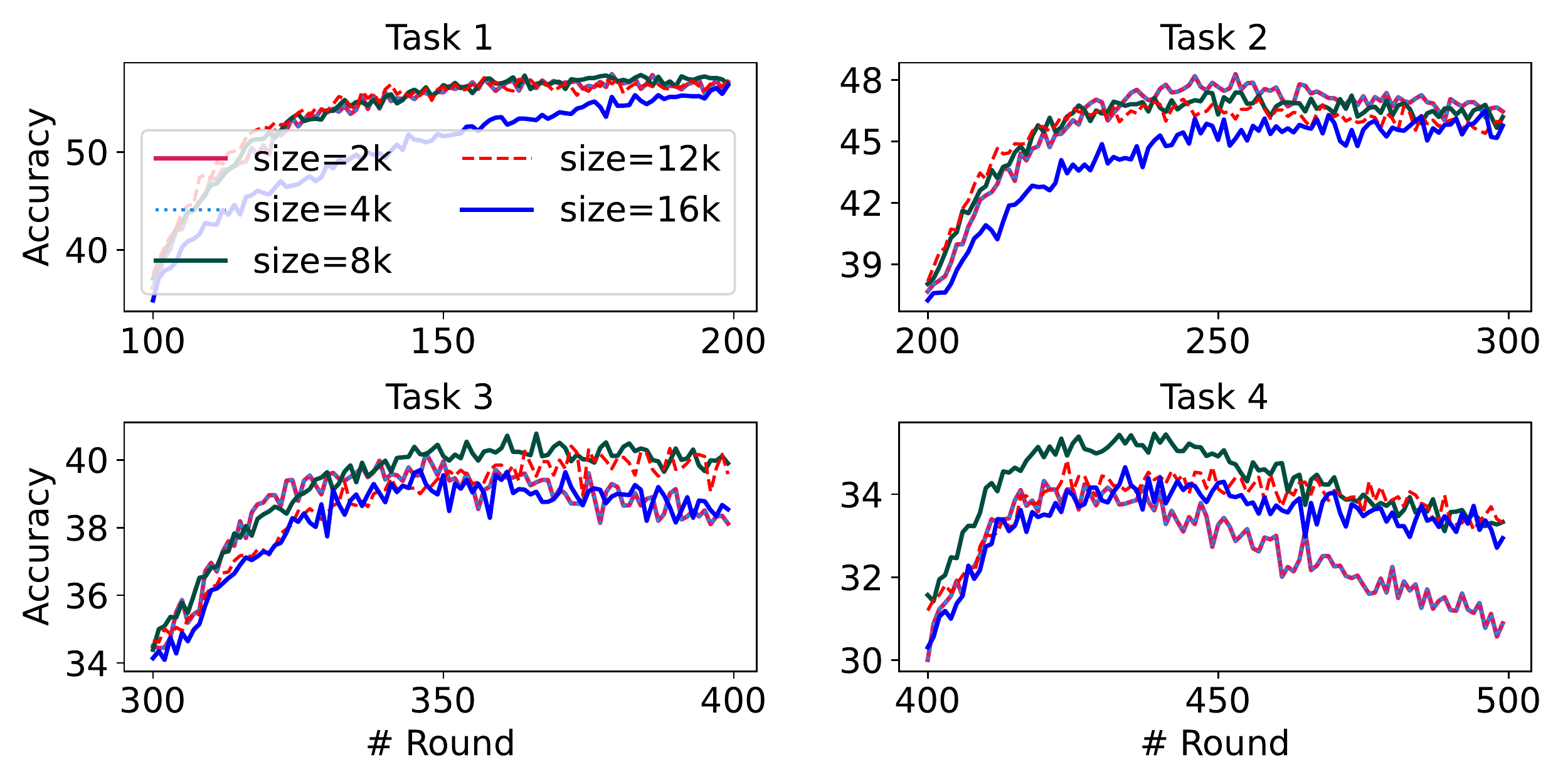}
\vspace{-4mm}
\caption{Accuracy for different size of synthetic dataset.}
\label{fig:size}
\end{figure}

\begin{figure}[t]
\centering
\includegraphics[width=8cm]{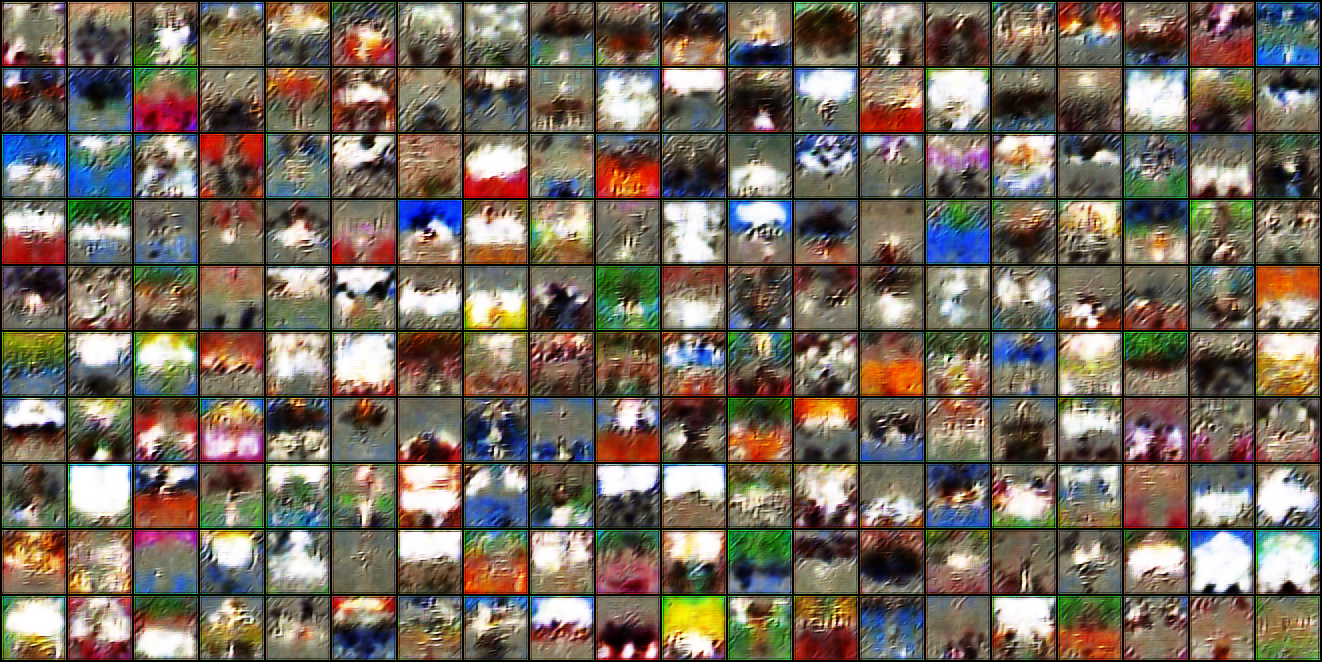}
\caption{Visualization of randomly synthesised data.}
\label{fig:cifar_vis}
\vspace{-4mm}
\end{figure}
\begin{figure}[t]
\centering
\includegraphics[width=8.5cm]{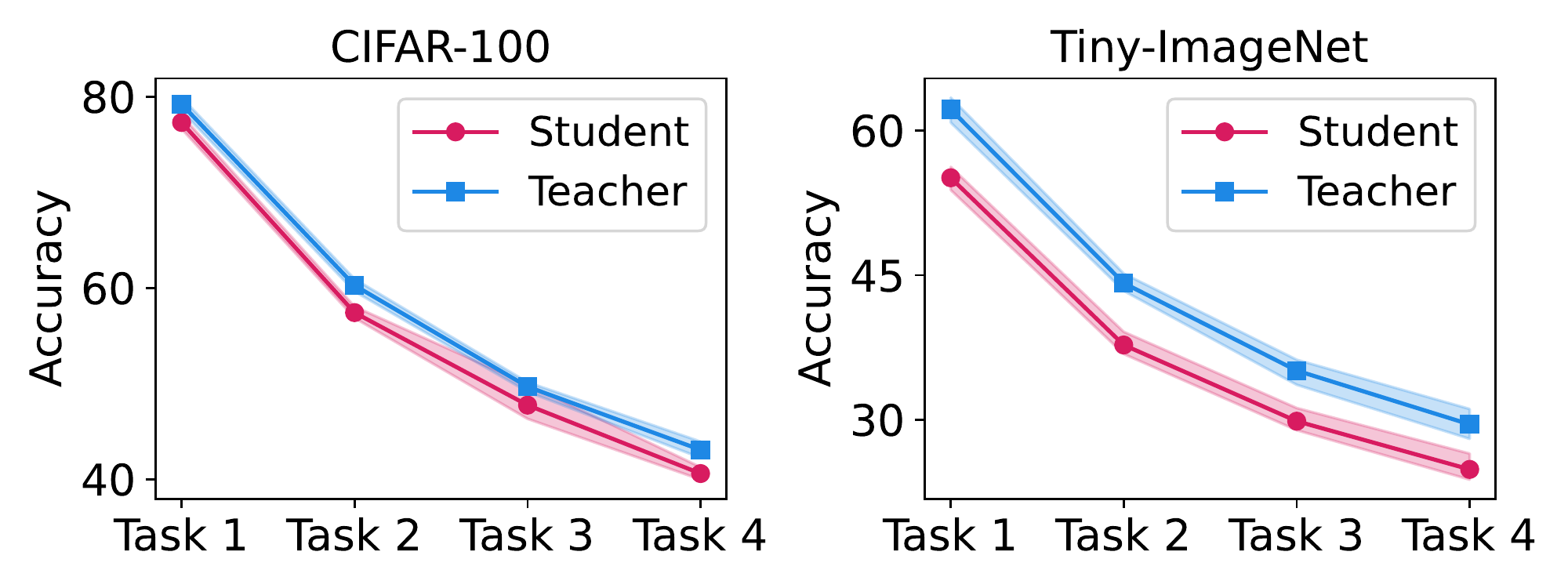}
\vspace{-3mm}
\caption{Distillation results for task $\{1,2,3,4\}$ when trained on 5 tasks.}
\label{fig:distill}
\vspace{-4mm}
\end{figure}

\paragraph{Memory of Synthetic Data.}
The optimal amount of synthetic data generated for the old task plays a crucial role in determining the final performance of our method. Insufficient synthetic data may not adequately facilitate knowledge transfer from the old tasks, while excessive data generation can lead to increased memory and communication costs. Therefore, determining an appropriate amount of synthetic data that strikes a balance between knowledge transfer and computational cost is critical for the effectiveness and efficiency of our approach. As shown in Figure~\ref{fig:size}, we test our method on CIFAR-100, which is divided into 5 tasks with different data 
sizes ranging from 2k to 16k. 
It can be observed that when the data volume is relatively small, such as 2k and 4k, the performance of the model is poor, especially when the data volume is 2k, the testing curve in task 4 shows a later decline. This is because when the data volume is too small, the model is unable to effectively learn knowledge from old tasks. Increasing the data volume to 8k can effectively alleviate the forgetting phenomenon and achieve good performance. However, continuously increasing the data volume to 12k and 16k do not result in significant improvement in the model's performance.
It is important to note that the size of the data volume alone does not guarantee the effectiveness of synthetic data in improving machine learning models. The quality and relevance of the synthetic data must also be carefully considered to ensure that it accurately represents the underlying distribution of the real-world data. 
\paragraph{Visualization on Synthetic Data.}
To demonstrate the effectiveness of the synthetic data, we present in Figure~\ref{fig:cifar_vis} the visualization results of the synthesized images generated by our method after the model learns the penultimate task on Tiny-ImageNet (CIFAR-100 in {Appendix}) for 5 tasks. These data have the potential to efficiently enable the student model to approach the performance of the teacher model rapidly. However, it is worth noting that these data exhibit visual dissimilarities from the actual training data.
\paragraph{Analysis on Distillation.} 
Based on the synthetic data, we distill the knowledge of the model trained on old tasks into a student model. The effect of distillation is further demonstrated in Figure~\ref{fig:distill}. It can be observed that even without accessing any 
private training data from clients, our method can quickly distill the student model on the server to approach the performance of the global model.

\section{Conclusion}
In conclusion, this paper introduces a novel method, TARGET (federat\textbf{T}ed cl\textbf{A}ss-continual lea\textbf{R}nin\textbf{G} via \textbf{E}xemplar-free dis\textbf{T}illation), to alleviate the catastrophic forgetting problem in Federated Class-Continual Learning (FCCL). 
Unlike all the previous methods, our proposed method leverages global knowledge, without 
requiring any additional datasets or data from previous tasks, making it ideal for privacy-sensitive scenarios. Extensive experimental results 
demonstrate the effectiveness of our proposed method in comparison to existing FCCL methods.

{\small
\bibliographystyle{ieee_fullname}
\bibliography{egbib}
}

\end{document}